\documentclass[pdflatex,sn-mathphys-num]{sn-jnl}

\usepackage{graphicx}%
\usepackage{multirow}%
\usepackage{amsmath,amssymb,amsfonts}%
\usepackage{amsthm}%
\usepackage{mathrsfs}%
\usepackage[title]{appendix}%
\usepackage{xcolor}%
\usepackage{textcomp}%
\usepackage{manyfoot}%
\usepackage{booktabs}%
\usepackage{algorithm}%
\usepackage{algorithmicx}%
\usepackage{algpseudocode}%
\usepackage{listings}%
\usepackage{pdflscape} %

\theoremstyle{thmstyleone}%
\theoremstyle{thmstyletwo}%

\theoremstyle{thmstylethree}%

\begin{document}

\title[LGM-Pose: A Lightweight Global Modeling Network for Real-time Human Pose Estimation]{LGM-Pose: A Lightweight Global Modeling Network for Real-time Human Pose Estimation}


\author[1,3]{\fnm{Biao} \sur{Guo}}

\author[1]{\fnm{Cong} \sur{Zhou}}

\author[1]{\fnm{Fangmin} \sur{Guo}}

\author[1]{\fnm{Xiaonan} \sur{Luo}}

\author[4]{\fnm{Guibo} \sur{Luo}}

\author*[1,2]{\fnm{Feng} \sur{Zhang}}\email{Curry@mails.guet.edu.cn}

\affil*[1]{\orgdiv{School of Computer Science and Information Security}, \orgname{Guilin University of Electronic Technology}, \orgaddress{\city{Guilin}, \postcode{541004}, \country{China}}}

\affil[2]{\orgname{Guangxi Metaverse Application Engineering Center}, \orgaddress{\city{Nanning}, \postcode{530000}, \country{China}}}

\affil[3]{\orgname{Guangxi Key Laboratory of Trusted Software}, \orgaddress{\city{Guilin}, \postcode{541004}, \country{China}}}

\affil[4]{\orgdiv{School of Electronic and Computer Engineering, Shenzhen Graduate School}, \orgname{Peking University}, \orgaddress{\city{Shenzhen}, \postcode{518000}, \country{China}}}

\abstract{Most of the current top-down multi-person pose estimation lightweight methods are based on multi-branch parallel pure CNN network architecture, which often struggle to capture the global context required for detecting semantically complex keypoints and are hindered by high latency due to their intricate and redundant structures. In this article, an approximate single-branch lightweight global modeling network (LGM-Pose) is proposed to address these challenges. In the network, a lightweight MobileViM Block is designed with a proposed Lightweight Attentional Representation Module (LARM), which integrates information within and between patches using the Non-Parametric Transformation Operation(NPT-Op) to extract global information. Additionally, a novel Shuffle-Integrated Fusion Module (SFusion) is introduced to effectively integrate multi-scale information, mitigating performance degradation often observed in single-branch structures. Experimental evaluations on the COCO and MPII datasets demonstrate that our approach not only reduces the number of parameters compared to existing mainstream lightweight methods but also achieves superior performance and faster processing speeds.}




\maketitle

\section{Introduction}\label{sec1}

Human Pose Estimation (HPE) aims to simultaneously detect key points of each human body in the input image. It serves as a fundamental technology for many computer vision tasks, such as human action/activity recognition\cite{1,2}, virtual animation modeling\cite{3}, human pose tracking\cite{4,5}, and motion analysis\cite{6,7}. In practical applications, these tasks often need to be deployed on resource-constrained devices. Consequently, the research of high-performance, lightweight HPE models has become a hot topic in this field. These models not only need to maintain high accuracy in pose estimation but also need to minimize computational resource consumption to meet the constraints of resource-limited terminal environments.

Currently, most lightweight top-down multi-person pose estimation methods employ the multi-branch parallel architecture of the High-Resolution Network(HRNet)\cite{8} to effectively integrate multi-scale features and enhance keypoint localization accuracy. For instance, methods such as Lite-HRNet\cite{9}, Dite-HRNet\cite{10}, and EDite-HRNet\cite{37} have further optimized the CNN blocks within this framework to reduce computational costs while achieving excellent performance. However, the multi-branch structure relies heavily on frequent upsampling and downsampling operations, as well as a large number of $1$ x $1$ convolutional operations. These operations increase computational complexity, consequently leading to relatively higher latency. Additionally, these lightweight methods are all built on pure CNNs, whose inherent limitations in network size and convolutional kernels restrict the model's ability to capture global information. This limitation causes difficulties in identifying or accurately locating semantically complex key points in the scene.

In recent years, the Transformer architecture \cite{11,12,13,14,15,16}, known for its excellent global learning representation capabilities, has been applied to human pose estimation methods, demonstrating significant key point detection abilities. Some models \cite{13,14,15,16} combine Transformers with CNNs, leveraging the spatial sensitivity of CNNs and the global information interaction capabilities of Transformers to achieve more robust pose estimation. However, the high computational cost of Transformers makes these methods challenging to apply directly in resource-constrained scenarios. MobileViT\cite{18} adopts a strategy of dividing feature maps into patches and calculating attention for different patches, thereby designing a lightweight attention model. Nonetheless, the Transformer component used to extract global information in MobileViT still accounts for considerable computational cost. To the best of our knowledge, it has not yet been directly applied to lightweight human pose estimation methods.

To address the issues mentioned above, this article proposes a lightweight global modeling network (LGM-Pose) for 2D multi-person pose estimation, as illustrated in Fig \ref{fig1}. The network adopts a single-branch structure to reduce complexity and incorporates a global information interaction module, named MobileViM. MobileViM enhances global context information acquisition while reducing model parameters and improving performance. Inspired by MobileViT, MobileViM interacts with spatial information of features through patches. However, unlike MobileViT, MobileViM utilizes a designed LARM module, which facilitates information exchange between and within patches through Non-Parametric Transformation Operation and MLPs (Multilayer Perceptrons), achieving global information fusion without the high computational cost of Transformer structures. Additionally, to address the performance degradation caused by the lack of multi-scale information in a single-branch structure, we designed a lightweight convolutional fusion method named SFusion(Shuffle-Integrated Fusion Module). SFusion employs shuffle and grouped convolutions to merge feature maps generated during the feature extraction phase with those output by deconvolution, minimizing computational cost while preserving channel connectivity. This further enhances pose estimation capability. Experiments on the MS-COCO\cite{19} and MPII\cite{20} datasets demonstrate that LGM-Pose outperforms current lightweight models with the fewest model parameters.

The main contributions of this article are as follows: (1) A lightweight global modeling network named LGM-Pose is proposed for multi-person pose estimation. This network is capable of capturing multi-scale and global information with a simple structure, achieving excellent performance among current representative models. (2) The MobileViM Block is designed, which efficiently establishes long-range dependencies and captures global information, thereby enhancing the ability to accurately locate complex keypoints. (3) The SFusion convolutional fusion module is designed, utilizing an effective feature fusion strategy to build multi-scale contextual relationships, further improving the network's pose estimation capabilities.

The remainder of this paper is organized as follows. Section 2 provides a comprehensive review of related work, including recent advancements in the field of human pose estimation and the limitations of existing methods. In Section 3, we present the proposed methodology, detailing the key components and theoretical foundations of our approach. Section 4 describes the experimental setup, including datasets, evaluation metrics, and an analysis of the experimental results. Finally, Section 5 concludes the paper by summarizing the proposed method and highlighting the performance improvements achieved.

\section{Related works}\label{sec2}

\subsection{Human pose estimation}
In recent years, with the development of deep convolutional neural networks (DCNN), significant progress has been made in human pose estimation\cite{8,21,22}. The mainstream methods are based on keypoint heatmap detection\cite{21,22,23,24,25}. Current human pose estimation methods utilizing heatmap detection can be divided into two frameworks: top-down and bottom-up. Bottom-up methods first independently detect all body keypoints\cite{26,27} and then assemble the detected joints to form multiple human poses. Although these methods perform well in real-time processing, they are not effective in handling varying human scales and crowded scenes. Conversely, top-down methods first detect all human bounding boxes in an image and then independently estimate the human pose within each bounding box\cite{9,10,13,14}. This approach standardizes all individual images obtained by the human detector to the same size, reducing the algorithm's sensitivity to changes in body size. Compared to bottom-up methods, the top-down approach yields higher prediction accuracy. 

This article focuses on human pose estimation research based on the top-down method. Due to the small size and extreme sensitivity of human keypoints, high-resolution convolutional networks have been proposed to improve prediction accuracy. For example, the SimpleBase \cite{21} model uses ResNet\cite{28} as the backbone network, connecting a three-layer deconvolution module after the last convolutional layer to gradually restore low-resolution features to high-resolution features. However, the final high-resolution feature is upsampled from the previous low-resolution features, leading to information loss. Additionally, HRNet employs high-resolution representations\cite{8} by progressively adding subnets from high to low resolution, forming more stages. The multi-resolution subnets are connected in parallel and repeatedly undergo multi-scale fusion, allowing each high-to-low resolution representation to continuously receive information from other parallel representations. Consequently, many works\cite{9,10,11} have widely used HRNet as the backbone network. These deep learning-based methods can accurately locate human keypoints. However, to obtain and maintain high-resolution feature maps, upsampling and deconvolution significantly increase the number of network parameters. Moreover, by deepening and widening the network and increasing the number of neurons, establishing attention mechanisms that focus on image spatial context information can enhance the model's learning ability. However, this complexity makes the network model bulky, and the efficiency of keypoint detection relatively low. Therefore, designing a lightweight network that balances performance with the limited resources of edge devices has become a major research trend.

\subsection{Lightweight Human Pose Estimation}

In recent years, lightweight and efficient models have become a major focus in human pose estimation. While traditional CNN-based methods have achieved remarkable success, their computational complexity and inefficiency in capturing global dependencies remain significant challenges. To address these issues, researchers have proposed various lightweight designs and attention mechanisms, each with its own advantages and limitations.

CNN-based lightweight models typically focus on reducing computational costs while maintaining performance. For instance, DSPNet\cite{39} introduces a depthwise separable upsampling module and a lightweight self-attention mechanism to enhance spatial context learning. Although effective, its design still leaves room for further optimization in terms of latency. Building on this, Lite-HRNet\cite{9} combines ShuffleNet's shuffle blocks with HRNet's high-resolution design, achieving stronger performance than MobileNet\cite{33}, ShuffleNet\cite{34}, and Small HRNet\cite{9}. It further reduces complexity by replacing costly pointwise ($1$ × $1$) convolutions with conditional channel weighting. However, despite its low parameter count, Lite-HRNet suffers from high latency, which limits its practicality. Similarly, Dite-HRNet\cite{10} improves accuracy by designing dynamic convolution kernels to capture multi-scale context information, but it does not address the latency issue, which remains a significant challenge for lightweight networks. Experiments\cite{29} have shown that low FLOPs do not guarantee low latency, highlighting the need for a more balanced approach. In contrast, LitePose\cite{27} demonstrates that a well-designed single-branch architecture can achieve lower latency and better performance than multi-branch designs. Inspired by this, our proposed network adopts an approximately single-branch design, minimizing redundant computations to enable efficient real-time inference on edge devices.

While CNNs excel at capturing local features, their ability to model global dependencies is limited. To address this, attention mechanisms, particularly those inspired by the Transformer architecture\cite{17}, have been introduced to human pose estimation. TransPose\cite{13} leverages the Transformer to predict keypoint locations based on heatmaps, enhancing the interpretability of human pose estimation. Similarly, TokenPose\cite{14} proposes a novel method based on token representation to learn constraint relationships and appearance cues from images. Although Transformer-based models have achieved competitive performance, their computational complexity increases significantly due to the self-attention mechanism, making them less suitable for resource-constrained devices.

To mitigate this issue, MobileViT\cite{18} proposes a lightweight attention module by dividing the modeling into local and global parts. It splits the feature map into multiple patches, applies the Transformer to the same pixel positions across different patches, and uses convolution to merge different pixel positions within the same patch for global information interaction. While this approach achieves significant performance in image classification, the Transformer used for global modeling in MobileViT remains a computational bottleneck. In our previous work\cite{31}, we demonstrated the feasibility of optimizing Transformer architecture with MLP\cite{30}. Building on this foundation, we conduct lightweight research on using MLP for global modeling and propose the LGM-Pose network, which structurally optimizes the multi-branch pure CNN network and introduces a lightweight global interaction module to improve pose estimation capabilities.

\section{Method}

This section proposes a simplified approach by designing an architecture that approximates a single-branch structure, as illustrated in Figure 1, aiming to reduce computational redundancy and achieve low latency. The MobileVim module is introduced, which partitions feature maps into patches and processes them through the Non-Parametric Transformation Operation, leveraging Multi-Layer Perceptron (MLP) blocks for feature interaction to capture global feature information. Additionally, the SFusion module is designed to integrate multi-scale features, thereby enhancing pose estimation performance. The detailed implementation will be elaborated in the following sections.

\begin{figure}[ht]
\centering
\includegraphics[width=\textwidth]{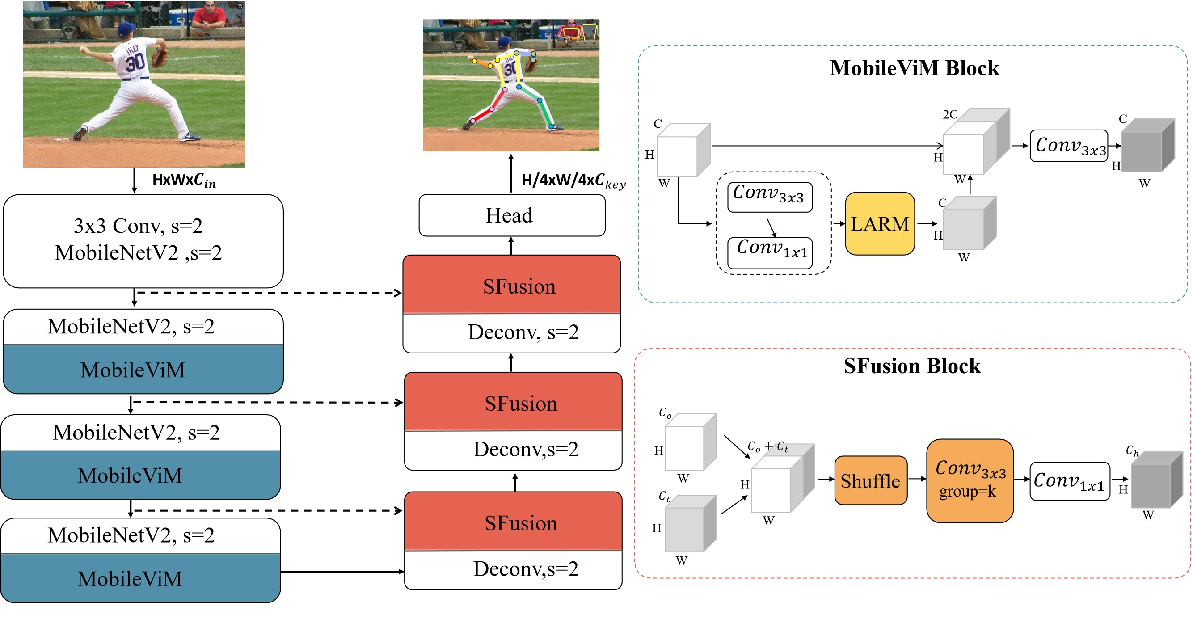}
\caption{The framework of LGM-Pose. LGM-Pose consists of MobileNetV2 Block, MobileViM Block, SFusion, Deconv, and a predict head. $C_{\text{key}}$ means the number of key points in the corresponding data set; "s" denotes the stride, which controls the sampling interval.}\label{fig1}

\end{figure}

\subsection{Architecture}
As illustrated in Fig \ref{fig1}, the proposed network architecture adopts a quasi-single-branch structure. The framework initially performs downsampling operations on the input feature maps, followed by upsampling processes. During this procedure, the network effectively extracts multi-scale feature information by integrating feature maps from corresponding downsampling and upsampling stages that share identical spatial dimensions. This design strategy ensures comprehensive multi-scale feature acquisition while minimizing redundant computations, thereby enhancing the overall network efficiency.

\subsection{MobileViM}

Inspired by MobileViT, MobileViM Block uses patches to interact with spatial information of features. However, unlike MobileViT, MobileViM does not use the computationally intensive Transformer module. Instead, it employs a designed LARM module to fuse global information within and between patches. As shown in Fig \ref{fig1}, the MobileViM Block first uses a standard $3$ x $3$ convolution to encode local information in the feature map, followed by a $1$ x $1$ convolution to learn the linear combination of input channels, projecting the tensor into a higher-dimensional space. The feature map is then input into the LARM module for global modeling. The LARM module uses Non-Parametric Transformation Operation to divide the feature map into patches, flatten, fold, and rotate them. Through the MLP Block, it facilitates global interactions among the same positional pixels across different patches and different positional pixels within the same patch, thereby achieving long-range perception capabilities. Due to the parameter-free nature of shape transformations and the minimal computational cost of the MLP Block, this module can significantly enhance performance with low computational overhead. Finally, the input features are convolved and fused with the globally modeled features to enhance the model's representation capabilities.

\subsubsection{LARM}

\begin{figure}[ht]
\centering
\includegraphics[width=\textwidth]{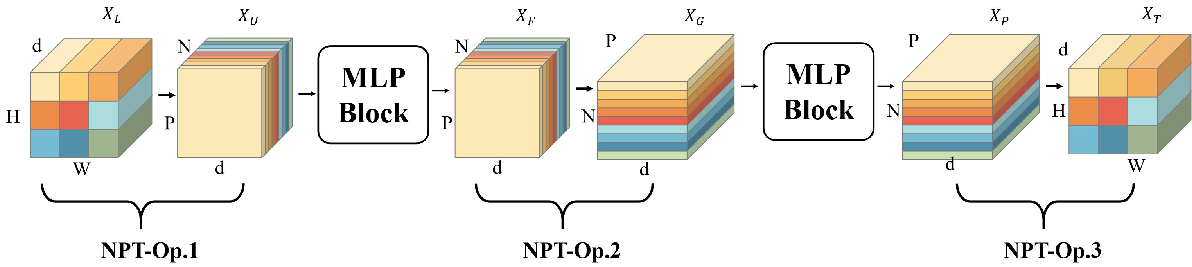}
\caption{LARM structure diagram}\label{fig2}
\end{figure}

The MobilieViT Block\cite{18} employs a Transformer to facilitate information exchange exclusively between patches. However, the Transformer component remains a computational bottleneck. This is primarily because the self-attention layer in Transformer models necessitates storing and processing a substantial amount of intermediate data, such as attention matrices, leading to a significant increase in memory requirements. To address this issue, we have designed the Lightweight Attentional Representation Module (LARM). This module employs only two MLP Blocks and the Non-Parametric Transformation Operation to achieve position-aware global modeling both between and within patches. The parameters of the MLP Block are comparable to those of the PerNorm and Feedforward components in the Transformer, effectively eliminating the computational bottleneck posed by the self-attention layer and significantly reducing the computational load.

As shown in Figure \ref{fig2}, the features from left to right are denoted as $\mathbf{X}_L, \mathbf{X}_U, \mathbf{X}_F, \mathbf{X}_G, \mathbf{X}_P, \mathbf{X}_T$. The process begins by decomposing $\mathbf{X}_L \in \mathbf{R}^{H \times W \times d}$ into $N$ non-overlapping flat planes $\mathbf{X}_L \in \mathbf{R}^{P \times N \times d}$. Here, $P = wh$, and $N = HW/P$ represents the number of patches, where $h \leq n$ and $w \leq n$ are the height and width of each patch, respectively. Then, transform $\mathbf{X}_L \in \mathbf{R}^{P \times N \times d}$ into $\mathbf{X}_U \in \mathbf{R}^{P \times d \times N}$, which is input into the MLP Block, resulting in an input dimension of $N$. The MLP Block can integrate pixels at the same position across different patches to produce tensor $\mathbf{X}_F \in \mathbf{R}^{P \times d \times N}$. The MLP Block is a two-layer perceptron, defined by the following equations:
\begin{equation}
\mathbf{x}_G = \text{Linear}\left(\text{GELU}\left(\text{Linear}\left(\text{LayerNorm}(\mathbf{x}_U)\right)\right)\right)
\end{equation}

The first linear layer maps the input into a higher-dimensional space to enhance nonlinear representation. The second linear layer then maps the high-dimensional information back to the original dimension. However, at this stage, the information exchange occurs only between different patches, and the pixels within the same patch remain independent. Therefore, a second round of information interaction is performed by transforming $\mathbf{X}_F \in \mathbf{R}^{P \times d \times N}$ into $\mathbf{X}_G \in \mathbf{R}^{N \times d \times P}$ and applying the MLP Block again to fuse information across different positions within the same patch. This allows for the interaction of information within the patch, resulting in a feature map $\mathbf{X}_P \in \mathbf{R}^{H \times W \times c}$ that incorporates global information. Finally, each patch is folded back to its initial form as  $\mathbf{X}_T \in \mathbf{R}^{H \times W \times d}$ .

\subsubsection{Non-Parametric Transformation Operation}
The Non-Parametric Transformation Operation is the primary reason for LARM's lightweight implementation, facilitating easy global information interaction. As shown in Fig \ref{fig2}, Non-Parametric Transformation Operation comprises three distinct operations: NPT-Op.1, NPT-Op.2, and NPT-Op.3. The NPT-Op.1 involves dividing the feature map into patches to obtain $\mathbf{R}^{P \times N \times d}$, and then flattening each patch into (1, N) and arranging them vertically, resulting in feature map dimension $\mathbf{R}^{P \times d \times N}$. During feature fusion, a nonlinear transformation is performed with N as the input, representing the pixels at the same position from different patches. The NPT-Op.2 flips $\mathbf{R}^{P \times d \times N}$ upwards to obtain feature map $\mathbf{R}^{N \times d \times P}$, where information interaction within the MLP Block module occurs between different pixels within the same patch. The NPT-Op.3 folds the feature map, post-information fusion, back to the initial feature map for the next stage of feature extraction. Despite a slight increase in latency, this parameter-free dimensional transformation, when combined with MLP, achieves excellent performance and is well-suited for lightweight networks.

\subsection{SFusion}
\begin{figure}[t]
\centering
\includegraphics[width=\textwidth]{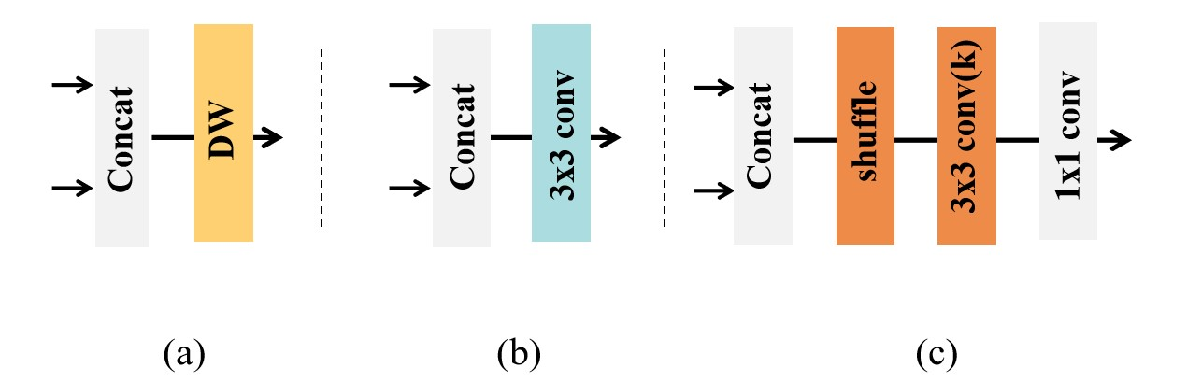}
\caption{(a) separable using depth, (b) conventional convolution, and (c) SFusion.}\label{fig3}
\end{figure}

The single-branch structure lacks multi-scale joint information, and deconvolution to generate high-resolution feature maps can result in significant loss of useful information. To address these issues, we propose a lightweight Shuffle-Integrated Fusion Module (SFusion). For the feature maps obtained after deconvolution, we concatenate them with the same resolution feature maps from the feature extraction stage, perform a shuffle operation, and then apply a group convolution. The shuffle operation and group convolution with a large number of groups allow information exchange between channels while maintaining a low parameter count, thereby enhancing model performance. As shown in Fig \ref{fig3}c, after concatenating the feature maps, we perform a shuffle operation followed by group convolution for feature fusion, where n is the number of shuffle groups and k is the number of group convolution groups, both simply set to 2 in our case. This configuration is chosen to balance model performance and computational efficiency, as increasing the number of groups would lead to higher computational complexity and memory usage, while reducing the number of groups might compromise the model's ability to learn diverse features. Compared to Fig \ref{fig3}a, where depthwise separable convolution for feature fusion loses inter-channel connections, and Fig \ref{fig3}b, where conventional  $3$ × $3$ convolution for feature fusion significantly increases the computational load, the SFusion module enhances channel connections with negligible computational cost, achieving optimal balance in the evaluation of computational cost and accuracy.

\section{Experiment}
In this section, the performance of the proposed LGM-Pose is evaluated using the publicly available MS COCO and MPII datasets. Additionally, ablation experiments are conducted on the MPII dataset to assess and analyze the effectiveness of the designed modules.

\subsection{Dataset \& Evaluation metric}
\subsubsection{Dataset}
The COCO dataset\cite{19} comprises over 200,000 images and 250,000 instances of people annotated with 17 key points. The COCO dataset includes three subsets: train 2017, val 2017, and test-dev 2017, containing 27K, 5K, and 20K images, respectively. In this study, the train2017 dataset is used for model training, while validation is conducted on the val2017 and test-dev2017 datasets. The MPII Human Pose dataset\cite{20} contains images with full-body pose annotations from real-world activities. It consists of 40K person samples with 16 joint labels. Of these, 12K samples are used as the test set, while the remaining samples serve as the training set.

The COCO and MPII datasets encompass a wide range of human poses and complex scenes, providing an effective evaluation of the model's accuracy and robustness in human keypoint detection.

\subsubsection{Evaluation metric}
The experiments employ the mAP (mean Average Precision) metric based on OKS on the COCO dataset, where OKS\cite{19} (Object Keypoint Similarity) defines the similarity between different human poses. The experimental results report the standard average precision and recall scores: AP(The average of AP scores across ten positions, OKS = 0.50, 0.55, ..., 0.90, 0.95), $\text{AP}^{50}$ (The AP at OKS of 0.50.), $\text{AP}^{70}$, $\text{AP}^M$, $\text{AP}^L$, and AR.

For the MPII dataset, the experiments utilize the standard metric PCKh @ 0.5\cite{20} (Percentage of Correct Keypoints at a 0.5 threshold) to assess performance.

\subsection{Implementation Details}
In this study, the experiments follow a two-stage top-down human pose estimation paradigm similar to SimpleBaseline \cite{21}. In this paradigm, a person detector first detects individual instances, and then keypoints are predicted. We use the human detector results provided by SimpleBaseline on the validation and test-dev sets.

In the experiments, the batch size for the GPU is set to 32, and Adam\cite{32} is used as the optimizer for the model. The initial learning rate is set to 1e-3, and it decays to 1e-4 and 1e-5 at the 200th and 240th epochs, respectively. The entire training process terminates after 260 epochs. The network is trained on a single NVIDIA A800 GPU. FPS measurements are conducted in both GPU and CPU environments, with the GPU environment utilizing a single Tesla M10 GPU and the CPU environment utilizing a single Intel(R) Xeon(R) Silver 4214 CPU @ 2.20GHz.

\subsection{Results}

\subsubsection{The Performance on MS COCO}

Since MS COCO is the most popular and representative dataset for human pose estimation tasks, we first compare the performance of LGM-Pose with other lightweight models that have demonstrated good results on the MS COCO dataset. During training, LGM-Pose is fed images with resolutions of $256$ x $192$ and $384$ x $288$. This allows for effective observation of LGM-Pose's superior performance compared to current lightweight models across different image input sizes.

\begin{sidewaystable}
\caption{\textbf{Comparisons of results on the COCO val2017 set}. P = pretrain the backbone on the ImageNet classification task. \textbf{Bold} indicates the best result and \underline{underline} indicates the second-best result. "-" indicates that the corresponding model's article does not provide the relevant code and numerical values}\label{tab1}
\begin{tabular*}{\textheight}{@{}lllllllllllll@{}}
\toprule%
Model & P & Input size & Params (M) & FLOPs (G) & AP & AP$^{50}$ & AP$^{75}$ & AP$^M$ & AP$^L$ & AR & CPU & GPU \\ 
\midrule
\multicolumn{13}{c}{\textbf{Large networks}} \\ 
\hline
Hourglass\cite{25} & N & 256x192 & 25.1 & 14.3 & 66.9 & - & - & - & - & - & - & - \\
CPN\cite{22} & Y & 256x192 & 27.0 & 6.2 & 68.6 & - & - & - & - & - & - & - \\
SimpleBaseline\cite{21} & Y & 256x192 & 34.0 & 8.9 & 70.4 & 88.6 & 78.3 & 67.1 & 77.2 & 76.3 & 7.8 & 26.2 \\
HRNet\cite{8} & N & 256x192 & 28.5 & 7.1 & 73.4 & 89.5 & 80.7 & 70.2 & 80.1 & 78.9 & 6.6 & 22.1 \\
TokenPose-B\cite{14} & N & 256x192 & 13.5 & 5.7 & 74.7 & 89.8 & 81.4 & 71.3 & 81.4 & 80.0 & 4.5 & 19.4 \\
MSFNet\cite{39} & N & 256x192 & - & 7.4 & 74.8 & 90.4 & 82.1 & 71.5 & 81.5 & 80.1 & - & - \\ 
\hline
\multicolumn{13}{c}{\textbf{Small networks}} \\ 
\hline
MobileNetV2 1\cite{33} & N & 256x192 & 9.6 & 1.4 & 64.6 & 87.4 & 72.3 & 61.1 & 71.2 & 70.7 & \underline{24.5} & \underline{74.0} \\
ShuffleNetV2 1\cite{34} & N & 256x192 & 7.6 & 1.2 & 59.9 & 85.4 & 66.3 & 56.6 & 66.2 & 66.4 & 20.3 & 71.2 \\
Small HRNet\cite{9} & N & 256x192 & 1.3 & 0.55 & 55.2 & 83.7 & 62.4 & 50.3 & 61.0 & 62.1 & 7.9 & 28.4 \\
Lite-HRNet-18\cite{9} & N & 256x192 & \textbf{1.1} &\textbf{0.2} & 64.8 & 86.7 & 73.0 & 62.1 & 70.5 & 71.2 & 7.4 & 27.2 \\
Lite-HRNet-30\cite{9} & N & 256x192 & 1.8 & 0.3 & 67.2 & 88.0 & 75.0 & 64.3 & 73.1 & 73.3 & 4.3 & 16.2 \\
Dite-HRNet-18\cite{10} & N & 256x192 & \textbf{1.1} & \textbf{0.2} & 65.9 & 87.3 & 74.0 & 63.2 & 71.6 & 72.1 & 5.5 & 9.4 \\
Dite-HRNet-30\cite{10} & N & 256x192 & 1.8 & 0.3 & 68.3 & 88.2 & \underline{76.2} & \underline{65.5} & \textbf{74.1} & \underline{74.2} & 3.3 & 9.4 \\
EDite-HRNet-s\cite{37} & N & 256x192 & 1.3 & \textbf{0.2} & 66.1 & 87.2 & 73.9 & 63.4 & 71.8 & 72.3 & - & - \\
LMFormer-L\cite{40} & N & 256x192 & 4.1 & 1.4 & \underline{68.9} & \underline{88.3} & \textbf{76.4} & - & - & \textbf{74.7} & - & - \\
Qi et al.\cite{41} & N & 256x192 & \underline{1.2} & \underline{0.28} & 65.9 & 87.7 & 74.1 & 62.3 & 70.6 & 71.3 & - & - \\
LGM-Pose(ours) & N & 256x192 & \textbf{1.1} & 0.6 & \textbf{69.3} & \textbf{89.5} & \underline{76.2} & \textbf{66.3} & \underline{73.5} & 73.7 & \textbf{33.1} & \textbf{80.1} \\ 
\hline
MobileNetV2 1\cite{33} & N & 384x288 & 9.6 & 3.3 & 67.3 & 87.9 & 74.3 & 62.8 & 74.7 & 72.9 & \underline{21.7} & \textbf{72.4} \\
ShuffleNetV2 1\cite{34} & N & 384x288 & 7.6 & 2.8 & 63.6 & 86.5 & 70.5 & 59.5 & 70.7 & 69.7 & 19.9 & \underline{70.2} \\
Small HRNet\cite{9} & N & 384x288 & 1.3 & 1.2 & 56.0 & 83.8 & 63.0 & 52.4 & 62.6 & 62.6 & 7.1 & 28.1 \\
Lite-HRNet-18\cite{9} & N & 384x288 & \textbf{1.1} & \textbf{0.4} & 67.6 & 87.8 & 75.0 & 64.5 & 73.7 & 73.7 & 6.7 & 27.2 \\
Lite-HRNet-30\cite{9} & N & 384x288 & 1.8 & 0.7 & 70.4 & 88.7 & 77.7 & 67.5 & 76.3 & 76.2 & 3.9 & 15.8 \\
Dite-HRNet-18\cite{10} & N & 384x288 & \textbf{1.1} & \textbf{0.4} & 69.0 & 88.0 & 76.0 & 65.5 & 75.5 & 75.0 & 4.9 & 16.3 \\
Dite-HRNet-30\cite{10} & N & 384x288 & 1.8 & 0.7 & \textbf{71.5} & 88.9 & \textbf{78.2} & \underline{68.2} & \textbf{77.7} & \textbf{77.2} & 2.9 & 9.6 \\
EDite-HRNet-s\cite{37} & N & 384x288 & 1.3 & 0.6 & 68.6 & 87.5 & 75.7 & 65.2 & 74.9 & 74.4 & - & - \\
LMFormer-L\cite{40} & N & 384x288 & 4.1 & 3.5 & 70.5 & 88.4 & 77.6 & - & - & 76.1 & - & - \\
Qi et al.\cite{41} & N & 384x288 & 1.9 & \underline{0.53} & 70.9 & \underline{89.1} & \underline{78.1} & 68.0 & 76.3 & \underline{76.5} & - & - \\
LGM-Pose(ours) & N & 384x288 & \underline{1.2} & 1.5 & \underline{71.0} & \textbf{90.5} & \underline{78.1} & \textbf{68.3} & \underline{77.4} & \underline{76.5} & \textbf{24.5} & 43.2 \\ 
\botrule
\end{tabular*}
\end{sidewaystable}

\textbf{COCO val set.} In Tables \ref{tab1}, the model performance is validated on the COCO val set.

At an input image of 256x192. Compared to large networks such as Hourglass\cite{25} and CPN\cite{22}, LGM-Pose achieves a comparable AP score with only 4\% of their parameter count. Although LGM-Pose's accuracy is not the highest when compared to HRNet\cite{8} and TokenPose-B\cite{14}, it offers significant advantages in terms of parameter count and latency, making it more suitable for deployment on resource-constrained edge devices. When compared to smaller networks, LGM-Pose achieves AP score improvements of 4.5 and 3.4 over LiteHRNet-18\cite{9} and DiteHRNet-18\cite{10}, respectively, while maintaining the same parameter count. Additionally, LGM-Pose increases FPS on CPU by 6x and 8x, and on GPU by 4x and 3x, respectively. This indicates that the redundancy of multi-resolution parallel architectures can lead to high latency, and that a well-designed single-branch structure can effectively address high latency issues while maintaining comparable performance. This also demonstrates the effectiveness of our designed modules for multi-person pose estimation. LGM-Pose has fewer parameters than EDite-HRNet-s\cite{37} but achieves a higher AP score. It also achieves nearly equivalent performance to LMFormer\cite{40} with only 27\% of the parameters. This indicates that LGM-Pose can obtain sufficient information while maintaining a small parameter count. The MobileViM Block captures long-range information without increasing network parameters to gather additional information. The table shows that the FPS of Dite-HRNet-18 is significantly lower than that of HRNet and the simple baseline, while LGM-Pose achieves 3 times the FPS of Dite-HRNet-18. This is because most current lightweight models rely on depthwise separable convolutions to reduce computational cost, which introduces significant latency during real-time inference. In contrast, our model reduces the use of depthwise separable modules and achieves global information fusion through feature map folding and rotation, enhancing feature extraction. As a result, our model maintains lightweight properties without a significant increase in inference latency, demonstrating the rationality of our design. The FPS values for EDite-HRNets and LMFormer-L are not provided due to the lack of publicly available code in their respective papers.

Compared to MobileNetV2\cite{33}, LGM-Pose uses only 12\% of the parameters while improving the AP by 4.0 points and achieving lower latency. Compared to ShuffleNetV2\cite{34}, LGM-Pose increases the AP by 8.7 points. When compared to Small HRNet\cite{9}, LGM-Pose's AP score is over 10 points higher. In terms of FPS, LGM-Pose's latency is significantly lower than these small networks. This demonstrates that our designed global interaction module and convolution fusion module can effectively alleviate the lack of global information capture capability and the absence of multi-scale features in pure CNN single-branch networks. Moreover, these designs are simple and effective, maintaining low latency. Compared to Dite-HRNet-30\cite{10}, LGM-Pose achieves an AP accuracy of 68.6 with only 61\% of the parameter count, an increase of 0.3 points, though the AR is slightly lower than that of Dite-HRNet-30. This indicates that high-resolution parallel networks can effectively combine multi-scale features for accurate keypoint localization. While adopting a single-branch structure inherently leads to some loss of scale sensitivity, well-designed convolution fusion and global modeling modules can significantly reduce the parameter count while maintaining comparable or even superior performance. Furthermore, LGM-Pose achieves the highest accuracy among current lightweight networks, with AP scores of 89.5 and 65.8 on datasets $\text{AP}^{50}$  and $\text{AP}^{M}$ , respectively. This indicates that the model has a stronger advantage in medium target pose estimation.

At an input image of 384x288. Compared to similarly parameterized lightweight networks like Lite-HRNet-18\cite{9} and Dite-HRNet-18\cite{10}, LGM-Pose achieves an AP score of 71.0, an improvement of 3.4 and 2.0 points, respectively. Although LGM-Pose's accuracy is slightly lower compared to the larger Dite-HRNet-30\cite{10}, it brings a substantial reduction in computational load with only a minor sacrifice in accuracy, demonstrating a strong advantage of our model. Compared to EDite-HRNet\cite{37} and LMformer\cite{40}, our model achieves higher performance with fewer parameters. This indicates that LGM-Pose combines the strengths of both single-branch and multi-branch structures, effectively capturing global and multi-scale features while reducing computational cost to enhance performance.

In the latency comparison of lightweight networks, the frame rate (FPS) of LGM-Pose on GPU decreases with increasing input image size. This indicates that LGM-Pose's sensitivity to input image size is related to the patch size setting. However, compared to multi-branch architectures like Lite-HRNet and Dite-HRNet, the FPS reduction of LGM-Pose with changes in image resolution is much less pronounced, highlighting the computational complexity advantages of a single-branch structure. Compared to lower latency networks such as MobileNet V2\cite{33} and ShuffleNet V2\cite{34}, LGM-Pose achieves a significant accuracy improvement while maintaining a low parameter count. This demonstrates that our designed lightweight global interaction module effectively achieves long-range modeling while keeping low complexity. Notably, LGM-Pose achieves the highest FPS on CPU, with minimal impact from input image size, making it particularly advantageous for deployment on resource-constrained edge devices and devices without GPUs. This results in a triple optimization of parameter count, accuracy, and real-time inference performance.

\begin{sidewaystable}
\caption{\textbf{Comparisons of results on the COCO test-dev2017 set}.\#Params and FLOPs are computed for pose estimation, and those for human detection are not included. \textbf{Bold} indicates the best result and \underline{underline} indicates the second-best result}\label{tab2}%
\begin{tabular}{@{}llllllllll@{}}
\toprule
Model  & Input size & Params (M) & FLOPs (G) & AP & AP$^{50}$ & AP$^{75}$ & AP$^M$ & AP$^L$ & AR \\ 
\hline
\multicolumn{10}{c}{\textbf{Large networks}} \\ 
\hline
SimpleBaseline\cite{21}  & 256x192 & 34.0 & 8.9 & 70.0 & 90.9 & 77.9 & 66.8 & 75.8 & 75.6  \\
CPN\cite{22}  & 384x288 & - & - & 72.1 & 91.4 & 80.0 & 68.7 & 77.2 & 78.5  \\
HRNet\cite{8}  & 384x288 & 28.5 & 16.0 & 74.9 & 92.5 & 82.8 & 71.3 & 80.9 & 80.1  \\
UDP\cite{35}  & 384x288 & 28.7 & 16.1 & 76.1 & 92.5 & 83.5 & 72.8 & 82.0 & 81.3  \\ 
DARK\cite{36}  & 384x288 & 63.6 & 32.9 & 76.2 & 92.5 & 83.6 & 72.5 & 82.4 & 81.1  \\
\hline
\multicolumn{10}{c}{\textbf{Small networks}} \\ 
\hline
MobileNetV2 1\cite{33}  & 384x288 & 9.8 & 3.3 & 66.8 & 90.0 & 74.0 & 62.6 & 73.3 & 72.3  \\
ShuffleNetV2 1\cite{34}  & 384x288 & 7.6 & 2.8 & 62.9 & 88.5 & 69.4 & 58.9 & 69.3 & 68.9  \\
Small HRNet\cite{9} & 384x288 & 1.3 & 1.2 & 55.2 & 85.8 & 61.4 & 51.7 & 61.2 & 61.5  \\
Lite-HRNet-18\cite{9}  & 384x288 & \textbf{1.1} &\textbf{0.4} & 66.9 & 89.4 & 74.4 & 64.0 & 72.2 & 72.6  \\
Lite-HRNet-30\cite{9}  & 384x288 & 1.8 & \underline{0.7} & 69.7 & 90.7 & 77.5 & 66.9 & 75.0 & 75.4  \\
Dite-HRNet-18\cite{10}  & 384x288 & \textbf{1.1} & \textbf{0.4} & 68.4 & 89.9 & 75.8 & 65.2 & 73.8 & 74.4 \\
Dite-HRNet-30\cite{10}  & 384x288 & 1.8 & \underline{0.7} & \textbf{70.6} & \underline{90.8} & 78.2 & \underline{67.4} & \textbf{76.1} & \underline{76.4} \\

LGM-Pose(ours)  & 384x288 & \underline{1.2} & 1.5 & \underline{70.4} & \textbf{91.2} & \underline{77.8} & \textbf{67.5} & \underline{75.4} & \underline{75.7}  \\ 
\botrule
\end{tabular}
\end{sidewaystable}

\textbf{COCO test-dev}. Tables \ref{tab2} shows the performance of LGM-Pose on the COCO test-dev dataset. When the input size is 384 x 288, LGM-Pose achieves an AP score of 70.4. Compared to Dite-HRNet-18[10], LGM-Pose achieves a 2.0-point increase in AP with an additional 0.1M parameters. Compared to Dite-HRNet-30, LGM-Pose achieves comparable performance with significantly fewer parameters. Additionally, LGM-Pose achieves the highest AP scores among lightweight network models, with $\text{AP}^{50}$ and $\text{AP}^{M}$ scores of 91.2 and 67.5. This demonstrates that LGM-Pose's use of the convolutional fusion module can compensate for the loss of useful information when generating high-resolution feature maps in single-branch networks, while effectively combining multi-scale features for superior pose estimation, especially in medium target estimation. Compared to large models, LGM-Pose achieves higher model accuracy than SimpleBaseline with only 3\% of the parameter count. Compared to larger models like HRNet[8], UDP[35], and DARK[36], although LGM-Pose's accuracy is not the highest, its computational cost is only a fraction of these large models. This indicates that even with multi-layer convolution stacking to expand the receptive field, large CNN models still struggle with the inherent limitation of the receptive field size. Our designed global interaction module starts from the feature map patches, transforming and interacting with them to achieve global information fusion at a low computational cost and complexity. For resource-constrained edge devices, LGM-Pose, built on MobileViM, is more competitive.

\subsubsection{The Performance on MPII}

\begin{table}[h!]
\caption{\textbf{Comparisons of results on the MPII val set}. \textbf{Bold} indicates the best result and \underline{underline} indicates the second-best result}\label{tab3}%
\begin{tabular}{@{}lllllll@{}}
\toprule
Model  & Input size & Params (M) & FLOPs (G) & PCKh & FPS-CPU & FPS-GPU  \\ 
\hline
\multicolumn{7}{c}{\textbf{Large networks}} \\ 
\hline
SimpleBaseline\cite{21}  & 256x256 & 34.0 & 8.9 & 88.5 & 7.8 & 26.2  \\
HRNet\cite{8}  & 256x256 & 28.5& 7.1 & 90.1 & 6.6 & 22.1  \\
TokenPose\cite{14}  & 256x256 & 21.4 & 9.1 & 90.2 & 4.5 & 19.4   \\ 
\hline
\multicolumn{7}{c}{\textbf{Small networks}} \\ 
\hline
MobileNetV2 1\cite{33}  & 256x256 & 9.6 & 1.97 & 85.4 & 24.3 & 69.7  \\
ShuffleNetV2 1\cite{34}  & 256x256 & 7.6 & 1.7 & 82.8 & 20.3 & \underline{72.2}  \\
Small HRNet\cite{9} & 256x256 & 1.3 & 0.72 & 80.2 & 7.3 & 28.3  \\
Lite-HRNet-18\cite{9}  & 256x256 & \textbf{1.1} &\textbf{0.27} & 86.1 & 6.8 & 27.1  \\
Lite-HRNet-30\cite{9}  & 256x256 & 1.8 & 0.42 & 87.0 & 4.0 & 15.3 \\
Dite-HRNet-18\cite{10}  & 256x256 & \textbf{1.1} & \textbf{0.27} & 87.0 & 5.1 & 16.0 \\
Dite-HRNet-30\cite{10}  & 256x256 & 1.8 & 0.44 & \underline{87.6} & 3.0 & 9.2 \\
EDite-HRNet-s\cite{37}  & 256x256 & 1.3 & \underline{0.3} & 86.8 & - & -  \\ 
LMFormer-L\cite{40}  & 256x256 & 4.1 & 1.9 & \underline{87.6} & - & -  \\ 
Qi et al.\cite{41}  & 256x256 & \underline{1.2} & 0.35 & 86.8 & - & -  \\ 
LGM-Pose(ours)  & 256x256 & \textbf{1.1} & 0.9 & \textbf{88.4} & \textbf{24.4} & \textbf{79.9}  \\ 
\botrule
\end{tabular}
\end{table}

As shown in Tables \ref{tab3}, the results compare LGM-Pose with other state-of-the-art lightweight networks on the MPII validation set. LGM-Pose achieves 88.4 PCK with 1.1M parameters, obtaining the highest accuracy among lightweight models. Compared to MobileNetV2\cite{33}, LGM-Pose improves the PCKh by 3 points. Compared to ShuffleNetV2\cite{34} and Small HRNet\cite{9}, LGM-Pose improves by 5.6 and 8.2 points, respectively. Compared to Dite-HRNet-30\cite{10}, LGM-Pose achieves better performance with 61\% of the parameter count, increasing the PCKh by 0.8 points. Compared to LMformer\cite{40}, LGM-Pose achieves higher accuracy with less than one-third of the parameters. This demonstrates that pure CNN networks lack the long-range dependency information required for human pose estimation tasks. Our designed global information interaction module enables lightweight global information fusion with positional information, achieving more robust pose representation capabilities. Additionally, compared to large networks, LGM-Pose achieves comparable scores with lower complexity, confirming the effectiveness of our global modeling module in capturing long-range context and mitigating the limited receptive field issue of small models for more precise keypoint localization.

In terms of real-time inference capability, LGM-Pose also achieves optimal performance. Lightweight networks based on multi-branch architectures typically exhibit higher latency due to architectural redundancy. LGM-Pose, designed with a single-branch architecture, achieves comparable FPS to MobileNetV2 and ShuffleNetV2 but with a lower parameter count and higher PCK. Compared to Lite-HRNet-30, LGM-Pose demonstrates superior parameter efficiency and accuracy, with FPS-CPU increasing by 8 times and FPS-GPU increasing by 5 times. Compared to Dite-HRNet-30, LGM-Pose's real-time inference speed improves nearly 8 times on both CPU and GPU. This proves that a well-designed single-branch architecture is more conducive to real-time inference on resource-constrained devices.

\subsection{Ablation Study}
 \subsubsection{MobileViM Block vs MobileViT Block}
\begin{table}[b]
\caption{MPII val set For ablation experiments using MobileViT Block and MobileViM Block.}\label{tab4}%
\begin{tabular}{@{}llllll@{}}
\toprule
\multicolumn{1}{c}{\textbf{Block}} & \textbf{Params} & \textbf{GFLOPs} & \textbf{PCKh} & \textbf{FPS-CPU(s)} & \textbf{FPS-GPU(s)} \\
\hline
MobileViT Block & 2.5M & 1.4 & 87.9 & 20.1 & 69.4 \\
MobileViM Block & 1.1M & 0.9 & 88.4 & 24.4 & 79.9 \\
\botrule
\end{tabular}
\end{table}

To validate the effectiveness of the designed MobileViM Block module, we constructed two networks with global interaction modules composed of either MobileViT Block[18] or MobileViM Block. As shown in Table \ref{tab4}, on the MPII validation set with an input image size of $256$ x $256$, the LGM-Pose built with MobileViM Block improves accuracy by 0.5 points compared to the network using MobileViT Block for global interaction, while reducing the parameter count by more than half. In terms of FLOPs, LGM-Pose reduces by 0.5 GFlops. Regarding latency, LGM-Pose increases by 4.3 FPS on CPU and 10.5 FPS on GPU.

 MobileViM re-designs the MobileViT Block by utilizing Non-Parametric Transformation Operation and simple MLP nonlinear transformations to achieve global information fusion. This approach eliminates the computational bottleneck of transformers, achieving improvements in both accuracy and latency. The LGM-Pose using MobileViM Block can achieve better performance with fewer parameters, making it suitable for integration into lightweight human pose estimation networks to enhance model performance.

  \subsubsection{Feature fusion is used by comparison}
\begin{table}[!ht]
\caption{Comparison of ablation experiments with different feature fusion modes on MPII val set. ``$g$'' represents the number of groups.}\label{tab5}%
\begin{tabular}{@{}llllll@{}}
\toprule
\multicolumn{1}{c}{\textbf{Method}} & \textbf{Params} & \textbf{GFLOPs} & \textbf{PCKh} & \textbf{FPS-CPU(s)} & \textbf{FPS-GPU(s)} \\
\midrule
None & 0.8M & 0.7 & 86.4 & 26.34 & 84.30 \\
Conv 3x3 & 2.1M & 1.5 & 88.6 & 24.78 & 80.63 \\
DW g=1 & 1.08M & 0.89 & 88.2 & 22.11 & 73.42 \\
DW g=2 & 1.1M & 0.90 & 88.3 & 24.21 & 79.12 \\
SFusion & 1.1M & 0.90 & 88.4 & 24.46 & 79.91 \\
\bottomrule
\end{tabular}
\end{table}

 To compare the impact of different feature fusion methods on model performance, we conducted experiments using various fusion techniques, including no feature fusion, concatenating feature maps followed by depthwise separable convolution, concatenating feature maps followed by convolution fusion, and the SFusion designed in this article. The input image size was $256$ x $256$, and training was performed on the MPII validation set. As shown in Tables \ref{tab5}, compared to no feature fusion, LGM-Pose achieved 2.0 PCK increase with only a slight increase in parameters. For a single-branch structure, lacking multi-scale information, SFusion can effectively combine multi-scale information to achieve a significant performance boost. Compared to using regular convolution, SFusion achieved comparable results with only half the computational cost. When compared to the depthwise separable feature fusion method, SFusion achieved an additional 0.2 increase in accuracy with almost negligible additional computational cost. This improvement is due to the fact that depthwise separable convolutions do not facilitate inter-channel communication. Our module uses group convolution, which maintains some channel connections without significant computational cost, thereby improving accuracy. Additionally, the shuffle operation enhances the model's robustness. Overall, SFusion achieves the best balance between accuracy and parameter count.

 To validate the effectiveness of the shuffle operation, we compared the results of the convolution fusion module with(SFuion) and without(DW g=2) the shuffle operation. As shown in Table \ref{tab5} the comparison, the model using the shuffle operation achieved a 0.1 increase in accuracy without any additional computational cost. This indicates that performing a shuffle operation after feature fusion can enhance inter-channel connections, improve feature representation, and achieve better performance.

 \subsection{Qualitative contrast}

\begin{figure}[ht]
\centering
\includegraphics[width=\textwidth]{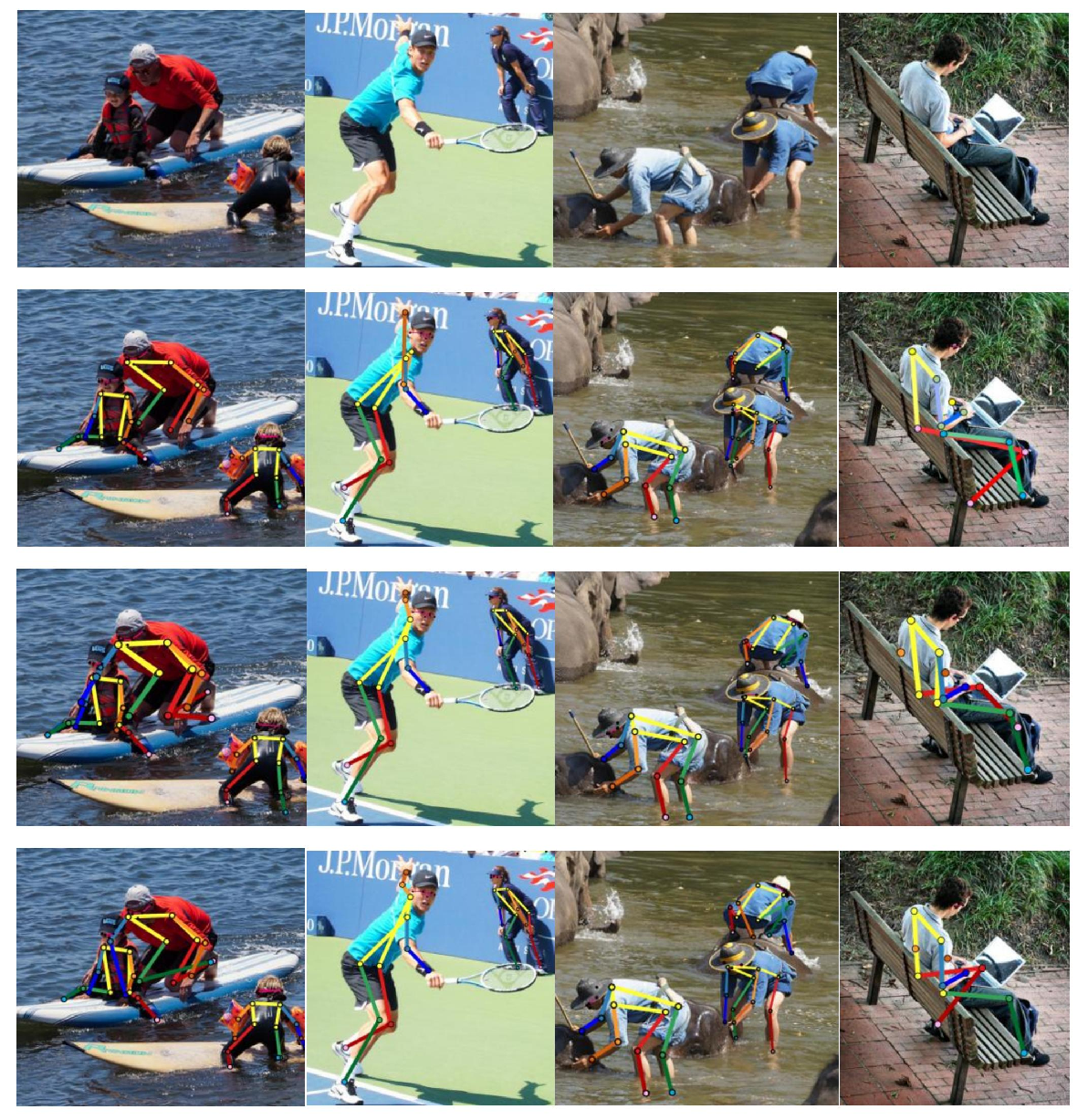}
\caption{Visualization images from the MS-COCO val set, with the first row being the original images, the second row showing the results of Lite-HRNet, the third row displaying the results of Dite-HRNet, and the fourth row presenting the results of LGM-Pose.}\label{fig4}
\end{figure}
In practical applications of multi-person pose estimation, challenges such as inter-person occlusion, self-occlusion, complex environments, and varying human scales often arise, making keypoint prediction difficult. To demonstrate the advantages of the LGM-Pose model, we present comparative illustrations in Fig. \ref{fig4}, showcasing the performance of LGM-Pose against other lightweight models in these challenging scenarios.

In the first column of images, severe inter-person and self-occlusion scenarios are presented. The second row (Lite-HRNet visualization) shows inaccurate leg joint localization for the child sitting on the boat and missed detection of the adult's leg joints. In the third row (Dite-HRNet visualization), although the child's leg joints are correctly detected, the adult's bent posture is still inaccurately localized. In the fourth row (LGM-Pose visualization), both the child's and the adult's poses are accurately detected and localized. This indicates that our model better understands the environment and human poses in the image, improving recognition in cases of occlusion and complex backgrounds.

In the second column of images, self-occlusion scenarios are presented. The second row (Lite-HRNet visualization) incorrectly positions the left arm, connecting it to the right elbow joint. The third row (Dite-HRNet visualization) inaccurately localizes the left elbow joint. In the fourth row (LGM-Pose visualization), all joints are more accurately positioned. This demonstrates that despite being a simple single-branch structure, our model, through the design of the multi-scale feature fusion module, effectively integrates features from low to high levels, enhancing the grasp of details and overall structure.

In the third column of images, varying human scales and severe self-occlusion scenarios are presented. The second row (Lite-HRNet visualization) shows incorrect localization of the right foot for the person in the front right, and the left arm and both legs for the person in the back. The third row (Dite-HRNet visualization) also shows incorrect localization of the right foot for the person in the front right and the left arm for the person in the back. In the fourth row (LGM-Pose visualization), all key points for all people in the image are correctly localized.

In the fourth column of images, complex environment scenarios are presented. The second row (Lite-HRNet visualization) and the third row (Dite-HRNet visualization) both show incorrect detection of the left arm and left leg. In the fourth row (LGM-Pose visualization), all key points are correctly detected. This demonstrates that our model, using the MobileViM Block, can perform context information interaction, achieving precise key point localization, particularly excelling in complex scenarios with severe occlusion.

\section{Conclusion}
This article presents LGM-Pose, a lightweight network offering an efficient solution for top-down human pose estimation on resource-constrained edge devices. The designed global module can be further extended to other image domains. LGM-Pose demonstrates excellent performance with low computational cost, low latency, and high accuracy, as evidenced by extensive experiments on the COCO and MPII human pose estimation datasets. In future work, we aim to optimize the model to address the relatively high GFLOPs, further enhancing its lightweight characteristics.

\section*{Acknowledgement}
This work is supported by the research project from Guangxi Metaverse Application Engineering Center (No.2023GMAEC003), the Science and Technology Base and Talent Project of Guangxi (No.AD21220097), the Project of Nanning Yongjiang Talent Plan (No.2019004, No.2023018), The Project of 2023 Innovation Platform and Talent Plan of Guilin Science and Technology Bureau, Research and Application Demonstration of Key Technologies for Immersive Experience in Cultural Tourism Metaverse (20232C19742623),  Guangxi Key Laboratory of Trusted Software (No.KX202057), and Guangxi Metaverse Application Engineering Center Demonstrates Innovative Applications of Large Language Models and Digital Avatar (20231C16737423), Guangxi University Young and Middle-aged Teachers Basic Research Ability Improvement Project (2021KY0202).

\bibliography{main}

\end{document}